# Unsupervised Feature Selection to Identify Important ICD-10 Codes for Machine Learning: A Case Study on a Coronary Artery Disease Patient Cohort

Peyman Ghasemi, MSc[1,2], Joon Lee, PhD[1,3,4,5]
[1] Data Intelligence for Health Lab, Cumming School of Medicine, University of Calgary, Calgary, AB, Canada; [2]Department of Biomedical Engineering, University of Calgary, Calgary, AB, Canada; [3]Department of Cardiac Sciences, Cumming School of Medicine, University of Calgary, Calgary, AB, Canada; [4]Department of Community Health Sciences, Cumming School of Medicine, University of Calgary, Calgary, AB, Canada; [5]Department of Preventive Medicine, School of Medicine, Kyung Hee University, Seoul, South Korea

**Abstract**
*The use of International Classification of Diseases (ICD) codes in healthcare presents a challenge in selecting relevant codes as features for machine learning models due to this system's large number of codes. In this study, we compared several unsupervised feature selection methods for an ICD code database of 49,075 coronary artery disease patients in Alberta, Canada. Specifically, we employed Laplacian Score, Unsupervised Feature Selection for Multi-Cluster Data, Autoencoder Inspired Unsupervised Feature Selection, Principal Feature Analysis, and Concrete Autoencoders with and without ICD tree weight adjustment to select the 100 best features from over 9,000 codes. We assessed the selected features based on their ability to reconstruct the initial feature space and predict 90-day mortality following discharge. Our findings revealed that the Concrete Autoencoder methods outperformed all other methods in both tasks. Furthermore, the weight adjustment in the Concrete Autoencoder method decreased the complexity of features.*

**Introduction**
Machine learning is increasingly being used in healthcare to analyze patient data and provide insights on improving health outcomes and quality of care[1]. With the rise of Electronic Health Data (EHD) and entering a large amount of data per each patient in hospitals, there are big opportunities to train machine learning models for a variety of applications, such as prediction/diagnosis of diseases, outcome prediction, and treatment planning[1,2]. EHD is a valuable source of information on a patient, containing details on their demographics, hospital visits, medical diagnoses, physiological measurements and treatments received [3]. However, despite the opportunities offered by these large datasets, there are challenges in terms of data quality, privacy concerns, and the complexity of medical conditions[1]. In terms of machine learning, EHD can include many irrelevant and redundant features that their direct use can lead to the "curse of dimensionality" as the high dimensionality of the data can make it more difficult to extract meaningful patterns and relationships[4]. Therefore, it is important to apply appropriate techniques for dimensionality reduction and feature engineering in order to address this challenge and improve the effectiveness of predictive models built from EHD.

Feature selection is one of the critical aspects of machine learning. It involves selecting a subset of relevant features that are most useful for predicting a target variable. In the case of medical data, these features could include patient demographics, medical history, laboratory test results, and diagnosis codes [3]. Feature selection is essential because it can help to improve the accuracy and performance of machine learning models by reducing the number of irrelevant or redundant features and avoiding overfitting[4]. Unsupervised feature selection is a type of feature selection method that is used when there is no target variable available to guide the selection of features. Unlike supervised feature selection, which chooses features predicting a certain target variable better, unsupervised feature selection methods rely on the intrinsic structure of the data to identify the most important features. This behavior helps the selected features to be unbiased and perform well when there is no labelled data and also it will reduce the risk of overfitting to a certain target variable and make the models robust to new outcome variables[5]. This is an important advantage in healthcare where collecting labelled data is usually difficult and also sometimes, we need to use the same data to predict multiple target variables.

Generally, there are three main categories of feature selection methods: filter, wrapper, and embedded methods. Filter methods utilize statistical tests like variance to rank individual features within a dataset and select the features that maximize the desired criteria, but they usually lack the ability to consider the interactions between features[6]. Wrapper methods, on the other hand, select features that optimize an objective function for a clustering algorithm. Therefore, these methods are generally specific to particular clustering algorithms and may not be suitable for use with other algorithms. Wrapper methods can detect potential relationships between features, but this often results in increased computational complexity[5]. Embedded methods, however, also take into account feature relationships, but generally do so more efficiently by incorporating feature selection into the learning phase of another algorithm. Lasso regularization is one of the well-known embedded methods that can be applied to a variety of machine learning models[6].

The 10th revision of the International Classification of Diseases (ICD-10) is a method of classifying diseases that was created by the World Health Organization and is used internationally[7]. It categorizes diseases based on their underlying cause, characteristics, symptoms, and location in the body, and uses codes to represent each disease. The ICD-10 system organizes thousands of codes in a hierarchical structure that includes chapters, sections, categories, and expansion codes. Within this structure, section codes and their corresponding chapter codes can be thought of as child-parent relationships, with each ICD-10 code serving as a node in the classification system. The same relationship applies to categories and sections, as well as expansion codes and categories. The high number of codes in this system is one of the important challenges of using them in machine learning applications[8]. It is worth noting that Canada has added or changed some codes in lower levels according to their healthcare system requirements (ICD-10-CA)[9].

In the current study, we used an administrative database containing ICD-10 codes related to admissions of patients with coronary artery disease (CAD) to acute care facilities to select the most informative codes about this cohort.

**Methods**

*Dataset and Preprocessing*
The Alberta Provincial Project for Outcome Assessment in Coronary Heart disease (APPROACH) Registry is one of the most comprehensive data repositories of coronary artery disease (CAD) management in the world, matching unique disease phenotypes with rich clinical information and relevant outcomes for patients in Alberta, Canada who have undergone diagnostic cardiac catheterization and/or revascularization procedures. These patients had undergone diagnostic angiography between January 2009 and March 2019 at one of the following three hospitals in Alberta: Foothills Medical Centre, University of Alberta Hospital, and Royal Alexandra Hospital. We excluded patients with ST-elevation myocardial infarction (STEMI) from the study in order to focus on non-emergency CAD.

Discharge Abstract Database (DAD) data between April 2004 and September 2022 for the above-mentioned patients were extracted from Alberta provincial health records. DAD contains information on the discharges or separations from all acute care facilities in Alberta. There were 49,075 unique patients included in this cohort (with at least one DAD record) and 273,910 DAD records were extracted.

Each DAD record can contain up to 25 ICD-10-CA codes. For each patient, we aggregated all ICD codes of hospital admissions within 3 months following the first admission date to make sure that chronic diseases have better representation. This decreased the number of records to 166,083. On average, each admission record in our DAD data set had 24.90 (SD=16.55) ICD codes. We one-hot encoded the ICD codes and their parent nodes for each record (e.g., if the code "*I251*" is present, "*I25*", "*I20-I25*", and "*Chapter IX*" are also encoded in the one-hot table). The final one-hot encoded table had 9,651 features ($N_{All} = 9,651$). Figure 1 shows the percentages of the 20 most common ICD-10-CA codes present in the processed dataset.

To validate the performance of the selected features in a real clinical problem, we pulled the mortality data of the patients enrolled in the cohort from the Vital Statistics Database (VSD) and matched them with the aggregated discharge records to determine 90-day post-hospital discharge mortality. 9,942 cases of 90-day mortality (20% mortality rate) were present in the dataset.

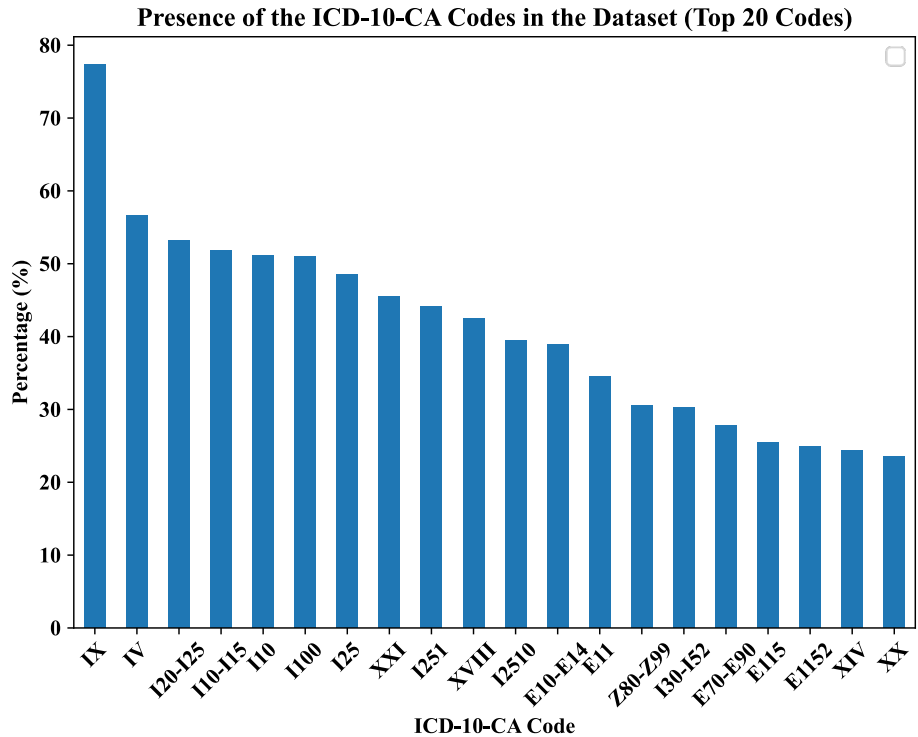

**Figure 1.** The percentages of the 20 most common ICD-10-CA codes present in the processed dataset

*Feature Selection*
The following unsupervised algorithms were utilized for feature selection:

- Concrete AutoEncoder (CAE)[6]: In this method, continuous relaxation of discrete random (Concrete) variables [10] and Gumbel-Softmax reparameterization trick can be used to construct a special layer in the neural network to transform discrete random variables into continuous ones, which allows for efficient computation and optimization using gradient-based methods. The reparameterization trick allows to use a softmax function in this layer which is differentiable unlike argmax function. This characteristic is useful for designing an autoencoder in which features are selected in the concrete layer (as the encoder) through the softmax operation and a common neural network (as the decoder) is used to reconstruct the main feature space out of the selected features. During the training a temperature parameter can be gradually decreased, allowing the concrete selector layer to try different features in the initial epochs and behave more like an argmax function in the last epochs to keep the best features. After training, we can use an argmax function on the weights to find the features passed to the neurons of the encoder layer. One of the major problems of this method is that it may converge to a solution in which some duplicate features are selected in some neurons (i.e., fewer than the desired number of features are selected).

- AutoEncoder Inspired Unsupervised Feature Selection (AEFS)[11]: This method combines autoencoder and group lasso tasks by applying an $L_{1,2}$ regularization on the weights of the autoencoder. The autoencoder in this method tries to map the inputs to a latent space and then reconstruct the inputs from that space. The $L_{1,2}$ regularization will optimize the weights to select a smaller number of features. The neural network structure of the autoencoder will enable the model to incorporate both linear and non-linear behavior of the data in the results.

- Principal Feature Analysis (PFA)[12]: This method selects features based on Principal Component Analysis (PCA). The most important features are selected by applying K-means clustering algorithm to the components of PCA and finding the features dominating each cluster.

- Unsupervised Feature Selection for Multi-Cluster Data (MCFS)[13]: This approach prioritizes the preservation of the multi-cluster structure of data. The algorithm involves constructing a nearest neighbor graph of the features, solving a sparse eigen-problem and an L1-regularized least squares problem to optimize feature selection.

- Laplacian Score (LS)[14]: The LS algorithm utilizes the nearest neighbor graph to capture the local structure of the data and computes the Laplacian score value for each feature. It is based on the observation that features that are close to each other usually belong to the same class.

We applied LS, AEFS, PFA, and CAE to a 67% training dataset of one-hot encoded features to select the best 100 features ($N_{Best} = 100$) with the following specifications (we chose this number based on preliminary experimentations):

For the AEFS method, we used a single hidden layer autoencoder and optimized the loss function described in [11] with $\alpha = 0.001$ as the trade-off parameter of the reconstruction loss and the regularization term and $\beta = 0.1$ as the penalty parameter for the weight decay regularization.

For the PFA method, we used incremental PCA, instead of the normal PCA in the original paper, with a batch size of $2N_{All}$ due to the high computational cost. We decomposed the data to $\frac{N_{All}}{2}$ components and then applied K-means clustering to find $N_{Best}$ clusters.

To employ LS and MCFS for feature selection, we utilized Euclidean distances between features to construct a nearest neighbor graph G based on the 5 nearest neighbors. For the LS method, we set the weights of the connected nodes of G to 1 assuming a large $t$ in the LS formulation. Then, we computed the Laplacian score for each feature and selected the top features with higher scores.

As the structure of the loss function allows us to prioritize some target variables, the CAE method was applied in two different ways of with/without adjusting weights for features. The reason for adjusting the weights is that since there are many correlated features in the ICD codes dataset, the model can choose one of them randomly [3]. Therefore, we applied the function in (equation 1) as the class weights of the features to the loss function of the model:

$$W_F = \frac{1}{1+d(F)} \qquad (1)$$

, where $W_F$ is the weight for feature F, and $d(F)$ is the depth of feature F as a node of the ICD-10-CA tree. This weight adjustment will enforce the model to give more importance to the features at the top of the ICD-10-CA tree and to generalize more in clinical settings.

We defined $N_{Best}$ neurons in the concrete selector layer and used a learning rate of 0.001, a batch size of 64, and 500 epochs. We also controlled the learning of the concrete selector layer by the temperature parameter that started from 20 and decreased to 0.01 exponentially. The decoder of the CAE was a feed-forward neural network with two hidden layers (with 64 neurons) and a sigmoid activation function for the output layer and a Leaky ReLU activation function for other layers.

*Evaluation of Features: Reconstruction of Initial Feature Space*
To evaluate the effectiveness of the selected features, we trained a simple feed-forward neural network model using the chosen features to reconstruct the initial feature space. The neural network consisted of two hidden layers, each with 64 neurons, and used Leaky ReLU activation functions, with a 10% dropout rate in the hidden layers, and a sigmoid activation function in the output layer. We trained the model using the same training set used in the feature selection step and evaluated its performance on the remaining 33% test set using binary cross entropy (BCE). We also calculated the accuracy of each feature selection method to determine which method produced the most accurate results. One of the challenges in comparing models with a large number of targets is that the accuracy values are inflated because most of the targets are heavily imbalanced (i.e., most of them were zeros), and the models were able to predict them easily. To tackle this problem, we used t-test analysis and compared the accuracy values of the classes with the accuracy of a model that simply outputs the mode of the training data for each class regardless of the input.

*Evaluation of Features: Prediction of 90-day Mortality*
To demonstrate the utility of using unsupervised feature selection methods in a supervised setting, we conducted a case study in which we used the selected features from each method to predict 90-day post-hospital discharge mortality. Since our dataset was highly imbalanced, with only ~6% of the admissions leading to 90-day mortality, we upsampled the minority class using random sampling to balance the dataset. We then trained a logistic regression model using the training set to predict the binary outcome variable and tested its performance on the test set.

**Results**
Table 1 shows the accuracies and BCEs of the models based on the selected features from each method. Table 2 shows the accuracy, F1-Score, Recall, and Precision metrics of the logistic regression models to predict 90-day mortality.

**Table** 1. Average Accuracy and Binary Cross Entropy Loss of different sets of selected features in reconstructing the initial feature space in a neural network structure.

| Feature Selection Method | CAE with weight adjustment | CAE without weight adjustment | AEFS | MCFS | PFA | LS | *Mode of Training Set* |
|---|---|---|---|---|---|---|---|
| Average Accuracy | 0.9988* | **0.9990*** | 0.9976 | 0.9984* | 0.9975 | 0.9989* | *0.9975* |
| BCE | 0.01848 | **0.01435** | 0.03566 | 0.2448 | 0.03832 | 0.01684 | *NA* |

* significantly different from the model that outputs the mode of each class ($p < 0.05$)

**Table 2.** Performance of different sets of selected features for predicting 90-day mortality in a logistic regression model.

| Feature Selection Method | CAE with weight adjustment | CAE without weight adjustment | AEFS | MCFS | PFA | LS |
|---|---|---|---|---|---|---|
| Accuracy | **0.801** | 0.797 | 0.547 | 0.723 | 0.502 | 0.749 |
| F1-Score | **0.788** | 0.7791 | 0.271 | 0.726 | 0.021 | 0.744 |
| Recall | 0.739 | **0.768** | 0.169 | 0.736 | 0.011 | 0.729 |
| Precision | **0.844** | 0.815 | 0.694 | 0.717 | 0.585 | 0.759 |

Both tables show that the CAE method was generally selecting better features compared to the other algorithms. Adjusting weight did not improve the performance of the feature selection in reconstructing the feature space, but it performed slightly better in predicting 90-day mortality. Table 3 shows the features selected by the best algorithm (CAE without adjusting the weights).

Figure 2 shows the log-scale histograms of initial feature space reconstruction accuracy in each ICD code for different feature selection methods. It shows that CAE without weight adjustment and LS were the best methods in terms of reconstructing the majority of the features with high accuracy. Other methods, despite having high average accuracy, performed poorly in reconstructing some of the features.

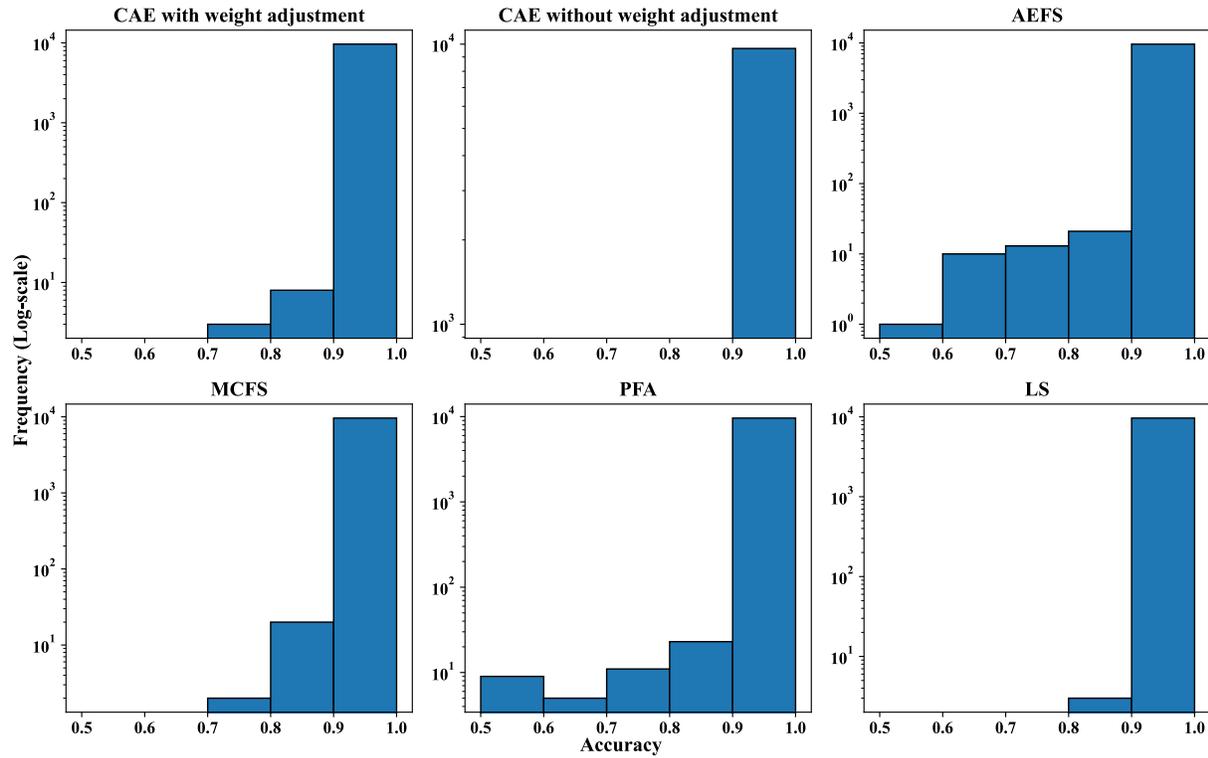

**Figure 2.** Log-scale histograms of initial feature space reconstruction accuracy in each ICD code for different feature selection methods.

**Table 3.** Selected Features by the best performing method (CAE without weight adjustment) with their chapters and descriptions[9]

| Chapter | Code | Description |
|---|---|---|
| I | A41 | Other sepsis |
| | B95-B98 | Bacterial, viral, and other infectious agents |
| II | C00-C75 | Malignant neoplasms stated or presumed to be primary, of specified sites, except of lymphoid, haematopoietic, and related tissue |
| | II | Neoplasms |
| | C30-C39 | Malignant neoplasms of respiratory and intrathoracic organs |
| | C76-C80 | Malignant neoplasms of ill-defined, secondary, and unspecified sites |
| III | III | Diseases of the blood and blood-forming organs and certain disorders involving the immune mechanism |
| | D60-D64 | Aplastic and other anaemias |
| IV | IV | Endocrine, nutritional, and metabolic diseases |
| | E10-E14 | Diabetes mellitus |
| | E11 | Type 2 diabetes mellitus |
| | E112 | Type 2 diabetes mellitus with renal complications |
| | E115 | Type 2 diabetes mellitus with peripheral circulatory complications |

|      | E1152    | Type 2 diabetes mellitus with certain circulatory complications |
|      | E65-E68  | Obesity and other hyperalimentation |
|      | E669     | Obesity, unspecified |
|      | E70-E90  | Metabolic disorders |
|      | E78      | Disorders of lipoprotein metabolism and other lipidaemias |
|      | E785     | Hyperlipidaemia, unspecified |
|      | E876     | Hypokalaemia |
|      | E877     | Fluid overload |
| V    | F03      | Unspecified dementia |
|      | F05      | Delirium, not induced by alcohol and other psychoactive substances |
|      | F30-F39  | Mood [affective] disorders |
| VI   | G40-G47  | Episodic and paroxysmal disorders |
|      | G50-G59  | Nerve, nerve root and plexus disorders |
| IX   | IX       | Diseases of the circulatory system |
|      | I10      | Essential (primary) hypertension |
|      | I20-I25  | Ischaemic heart diseases |
|      | I214     | Acute subendocardial myocardial infarction |
|      | I251     | Atherosclerotic heart disease |
|      | I2519    | Atherosclerotic heart disease of unspecified type of vessel, native or graft |
|      | I252     | Old myocardial infarction |
|      | I30-I52  | Other forms of heart disease |
|      | I442     | Atrioventricular block, complete |
|      | I48      | Atrial fibrillation and flutter |
|      | I4890    | Atrial fibrillation, unspecified |
|      | I50      | Heart failure |
|      | I60-I69  | Cerebrovascular diseases |
| X    | X        | Diseases of the respiratory system |
|      | J09-J18  | Influenza and pneumonia |
|      | J189     | Pneumonia, unspecified |
|      | J44      | Other chronic obstructive pulmonary disease |
|      | J459     | Asthma, unspecified |
|      | J95-J99  | Other diseases of the respiratory system |
| XI   | XI       | Diseases of the digestive system |
|      | K21      | Gastro-oesophageal reflux disease |
|      | K55-K64  | Other diseases of intestines |
|      | K65-K67  | Diseases of peritoneum |
|      | K800     | Calculus of gallbladder with acute cholecystitis |
| XII  | XII      | Diseases of the skin and subcutaneous tissue |
| XIII | M00-M25 | Arthropathies |
| | XIII | Diseases of the musculoskeletal system and connective tissue |
| | M25 | Other joint disorders, not elsewhere classified |
| XIV | XIV | Diseases of the genitourinary system |
| | N17 | Acute renal failure |
| | N179 | Acute renal failure, unspecified |
| | N18 | Chronic kidney disease |
| | N39 | Other disorders of urinary system |
| | N80-N98 | Noninflammatory disorders of female genital tract |
| XVIII | XVIII | Symptoms, signs and abnormal clinical and laboratory findings, not elsewhere classified |
| | R06 | Abnormalities of breathing |
| | R10-R19 | Symptoms and signs involving the digestive system and abdomen |
| | R25-R29 | Symptoms and signs involving the nervous and musculoskeletal systems |
| | R47-R49 | Symptoms and signs involving speech and voice |
| | R90-R94 | Abnormal findings on diagnostic imaging and in function studies, without diagnosis |
| | R94 | Abnormal results of function studies |
| | R943 | Abnormal results of cardiovascular function studies |
| XIX | XIX | Injury, poisoning and certain other consequences of external causes |
| | S20-S29 | Injuries to the thorax |
| XX | V01-X59 | Accidents |
| | XX | External causes of morbidity and mortality |
| | W00-W19 | Falls |
| | Y40-Y84 | Complications of medical and surgical care |
| | Y83 | Surgical operation and other surgical procedures as the cause of abnormal reaction of the patient, or of later complication, without mention of misadventure at the time of the procedure |
| XXI | Z22 | Carrier of infectious disease |
| | Z40-Z54 | Persons encountering health services for specific procedures and health care |
| | Z720 | Tobacco use |
| | Z80-Z99 | Persons with potential health hazards related to family and personal history and certain conditions influencing health status |
| | Z82 | Family history of certain disabilities and chronic diseases leading to disablement |
| | Z85 | Personal history of malignant neoplasm |
| | Z911 | Personal history of noncompliance with medical treatment and regimen |
| | Z921 | Personal history of long-term (current) use of anticoagulants |
| | Z94 | Transplanted organ and tissue status |
| | Z95 | Presence of cardiac and vascular implants and grafts |

## Discussion

The high dimensionality of ICD code databases necessitates the use of dimensionality reduction techniques to feed the data into machine learning models. Due to interpretability concerns in the health domain, selecting distinct features (rather than transforming them into new features) is an essential step in reducing dimensions. In this study, we demonstrated that the CAE method has the best capability of selecting the most informative ICD codes in an unsupervised setting. Using a clinical outcome as a case study, we also demonstrated that ICD code features selected by the CAE method can predict the outcome variable with better accuracy compared to other methods in the study, even though they were derived from an unsupervised setting in the absence of the target variable. This indicates that the selected features can be considered unbiased towards a specific target variable and explain the phenomenon appropriately. We also showed that AEFS and PFA methods did not select high-quality features in our dataset and are not suitable for both tasks of reconstructing the feature space and predicting 90-day mortality. LS and MCFS, however, showed better performance in both tasks (slightly lower than CAE).

MCFS, PFA, and LS had multiple matrix operations that made them computationally expensive. Considering our large-scale, high dimensional dataset, these algorithms were not possible to run on a normal computer and we had to optimize the operations for an advanced computing cluster with a high number of CPUs. AEFS and CAE, however, had the advantage of utilizing GPUs for training the neural networks and were faster.

Previous studies typically selected ICD codes as machine learning model features based on expert opinions or the presence of high-level codes (e.g., categories or chapters), or a combination of both[15,16]. To the best of our knowledge, only one study [3] attempted to offer a sophisticated feature selection method using tree-lasso regularization for ICD code datasets, but it was in a supervised setting that required an outcome variable. Our study provides a general tool for health researchers to filter out the most informative ICD codes without biasing the study towards a specific outcome variable. We also introduced a unique target weight adjustment function to the CAE model to guide the model to select higher levels of the ICD table (sum of feature depth of 140 in ICD-10-CA tree) compared to the model without adjustment (sum of feature depth of 148).

One of the limitations of this study was the incapability of the CAE method to select an exact number of desired features. Since the neurons in the concrete selector layer work independently, there is a possibility of selecting duplicate features. Therefore, the number of final selected features can be less than the desired number. Although it indicates that the decoder model is still capable of reconstructing the initial feature space with a smaller number of features, some researchers may prefer to have an exact number of features they desire for their models. One previous study[17] has used a special regularization term in the training step to enforce the model not to select duplicate features. This method can be investigated for the ICD codes in the future.

Another limitation was that we only used one dataset of a specific disease cohort to choose the features. Therefore, the selected features in this study may not be suitable in general. Furthermore, we selected the 100 best features but other data sets or patient cohorts may require a different number of features. Future studies may investigate the impact of the number of features on the results.

## Conclusion

In this study, we investigated five different methods for selecting the best feature in ICD code datasets in an unsupervised setting. We demonstrated that the CAE method can select better features representing the whole dataset and is useful in further machine learning studies. We also introduced weight adjustment of the CAE method for ICD code datasets that can be useful in the generalizability and interpretability of the models given that it prioritizes selecting high-level definitions of diseases.

## Acknowledgements

This study was supported by a Libin Cardiovascular Institute PhD Graduate Scholarship and a Project Grant from the Canadian Institutes of Health Research (PJT 178027).

## References

1. Jensen PB, Jensen LJ, Brunak S. Mining electronic health records: towards better research applications and clinical care. Nat Rev Genet. 2012 Jun;13(6):395–405.

2. Yu C, Liu J, Nemati S, Yin G. Reinforcement Learning in Healthcare: A Survey. ACM Comput Surv. 2023 Jan 31;55(1):1–36.

3. Kamkar I, Gupta SK, Phung D, Venkatesh S. Stable feature selection for clinical prediction: Exploiting ICD tree structure using Tree-Lasso. Journal of Biomedical Informatics. 2015 Feb 1;53:277–90.

4. Berisha V, Krantsevich C, Hahn PR, Hahn S, Dasarathy G, Turaga P, et al. Digital medicine and the curse of dimensionality. npj Digit Med. 2021 Oct 28;4(1):1–8.

5. Solorio-Fernández S, Carrasco-Ochoa JA, Martínez-Trinidad JFco. A review of unsupervised feature selection methods. Artif Intell Rev. 2020 Feb 1;53(2):907–48.

6. Abid A, Balin MF, Zou J. Concrete Autoencoders for Differentiable Feature Selection and Reconstruction [Internet]. arXiv; 2019 [cited 2023 Feb 25]. Available from: http://arxiv.org/abs/1901.09346

7. Organization WH. International Statistical Classification of Diseases and Related Health Problems: Alphabetical index. World Health Organization; 2004. 824 p.

8. Yan C, Fu X, Liu X, Zhang Y, Gao Y, Wu J, et al. A survey of automated International Classification of Diseases coding: development, challenges, and applications. Intelligent Medicine. 2022 Aug 1;2(3):161–73.

9. International Statistical Classification of Diseases and Related Health Problems (ICD-10-CA), Tenth Revision, Canada - Tabular List. Vol. 1. Ottawa: Canadian Institute for Health Information; 2015.

10. Maddison CJ, Mnih A, Teh YW. The Concrete Distribution: A Continuous Relaxation of Discrete Random Variables [Internet]. arXiv; 2017 [cited 2023 Feb 25]. Available from: http://arxiv.org/abs/1611.00712

11. Han K, Wang Y, Zhang C, Li C, Xu C. Autoencoder Inspired Unsupervised Feature Selection. In: 2018 IEEE International Conference on Acoustics, Speech and Signal Processing (ICASSP). 2018. p. 2941–5.

12. Lu Y, Cohen I, Zhou XS, Tian Q. Feature selection using principal feature analysis. In: Proceedings of the 15th ACM international conference on Multimedia. New York, NY, USA: Association for Computing Machinery; 2007. p. 301–4. (MM '07).

13. Cai D, Zhang C, He X. Unsupervised feature selection for multi-cluster data. In: Proceedings of the 16th ACM SIGKDD international conference on Knowledge discovery and data mining. New York, NY, USA: Association for Computing Machinery; 2010. p. 333–42. (KDD '10).

14. He X, Cai D, Niyogi P. Laplacian Score for Feature Selection. In: Advances in Neural Information Processing Systems. MIT Press; 2005.

15. Jamian L, Wheless L, Crofford LJ, Barnado A. Rule-based and machine learning algorithms identify patients with systemic sclerosis accurately in the electronic health record. Arthritis Research & Therapy. 2019 Dec 30;21(1):305.

16. Lucini FR, Stelfox HT, Lee J. Deep Learning–Based Recurrent Delirium Prediction in Critically Ill Patients. Critical Care Medicine. :10.1097/CCM.0000000000005789.

17. Strypsteen T, Bertrand A. End-to-end learnable EEG channel selection for deep neural networks with Gumbel-softmax. J Neural Eng. 2021 Jul;18(4):0460a9.